\documentclass[letterpaper, 10 pt, journal, twoside]{IEEEtran}

\usepackage{graphicx}
\usepackage{amsmath}
\usepackage{algorithm}
\usepackage{algpseudocode}
\usepackage{xcolor}
\usepackage{soul}
\usepackage{stfloats}
\usepackage{amsfonts} 
\usepackage{amssymb}
\usepackage{tabularx}
\usepackage{booktabs}
\usepackage{comment}
\usepackage{subcaption}
\usepackage{xspace}
\renewcommand{\baselinestretch}{1}
\usepackage{xcolor}
\usepackage{comment}
\usepackage{hyperref}
\newcommand\edit[1]{{\color{black}#1}}
\newcommand\markred[1]{{\color{black}#1}}
\newcommand\markgreen[1]{{\color{teal}#1}}
\usepackage{indentfirst}
\usepackage{todonotes}
\usepackage{tkz-graph}
\usetikzlibrary{automata,positioning}
\usepackage{tikz}
\usetikzlibrary{positioning}
\usepackage[nolist,nohyperlinks]{acronym}
\def\identitymat{\mathbf{I}}

\def\occ{o}
\def\distfield{{\hat{d}}}

\def\revert{r}

\def\lengthscale{l}

\def\abscissavec{\mathbf{x}}
\def\abscissamat{\mathbf{X}}

\newcommand{\kernel}[3]{{{k}_{#1}\left(#2,#3\right)}}

\newcommand{\kernelmatraw}[1]{{\mathbf{K}_{#1}}}

\newcommand{\kernelvecraw}[1]{{\mathbf{k}_{#1}}}

\usepackage{pifont}
\usepackage{bbding}

\ifCLASSINFOpdf
\else
\fi

\begin{document}

\title{\LARGE \bf
VDB-GPDF: \\
Online Gaussian Process Distance Field with VDB Structure
}

\author{Lan Wu$^{1}$, Cedric Le Gentil$^{1,}$$^{2}$, and Teresa Vidal-Calleja$^{1}$%
\thanks{$^{1}$Robotics Institute, University of Technology Sydney, Ultimo, NSW 2007, Australia. Corresponding author: {\tt\footnotesize Lan.Wu-2@uts.edu.au}}
\thanks{$^{2}$Autonomous Space Robotics Lab, University of Toronto, Ontario, Canada.}
\thanks{This work was supported by ARIA Research and the Australian Government via the Department of Industry, Science, and Resources CRC-P program (CRCPXI000007) and the Australian Research Council Discovery Project under Grant DP210101336.}
\thanks{This work has been submitted to the IEEE for possible publication. Copyright may be transferred without notice, after which this version may no longer be accessible.}%
}

\maketitle

\begin{abstract}
Robots reason about the environment through dedicated representations. Popular choices for dense representations exploit Truncated Signed Distance Functions (TSDF) and Octree data structures. However, \edit{TSDF provides a projective or non-projective signed distance obtained directly from depth measurements that overestimate the Euclidean distance.} Octrees, despite being memory efficient, require tree traversal and can lead to increased runtime in large scenarios. Other representations based on the Gaussian Process (GP) distance fields are appealing due to their probabilistic and continuous nature, but the computational complexity is a concern. 
In this paper, we present an online efficient mapping framework that seamlessly couples GP distance fields and the fast-access OpenVDB data structure. 
The key aspect is a latent Local GP Signed Distance Field (L-GPDF) contained in a local VDB structure that allows fast queries of the Euclidean distance, surface properties and their uncertainties for arbitrary points in the field of view. Probabilistic fusion is then performed by merging the inferred values of these points into a global VDB structure that is efficiently maintained over time. After fusion, the surface mesh is recovered, and a global GP Signed Distance Field (G-GPDF) is generated and made available for downstream applications to query accurate distance and gradients. 
A comparison with the state-of-the-art frameworks shows superior efficiency and accuracy of the inferred distance field and comparable reconstruction performance. {\tt \url{https://github.com/UTS-RI/VDB_GPDF}}
\end{abstract}


\begin{figure}[ht]
  \centering
  \resizebox{\linewidth}{!}{
  \subfloat[ \label{teaser1}]{\includegraphics[height=4cm]{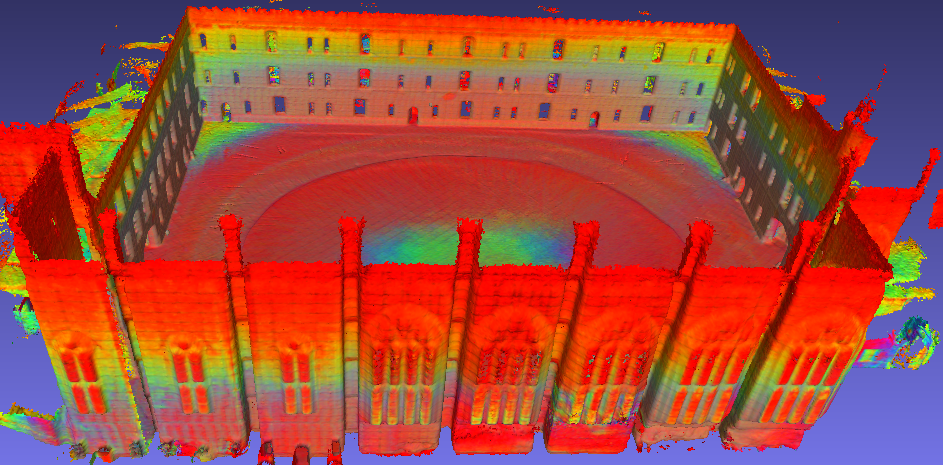}}}
  \resizebox{\linewidth}{!}{
  \subfloat[ \label{teaser2}]{\includegraphics[height=3.5cm]{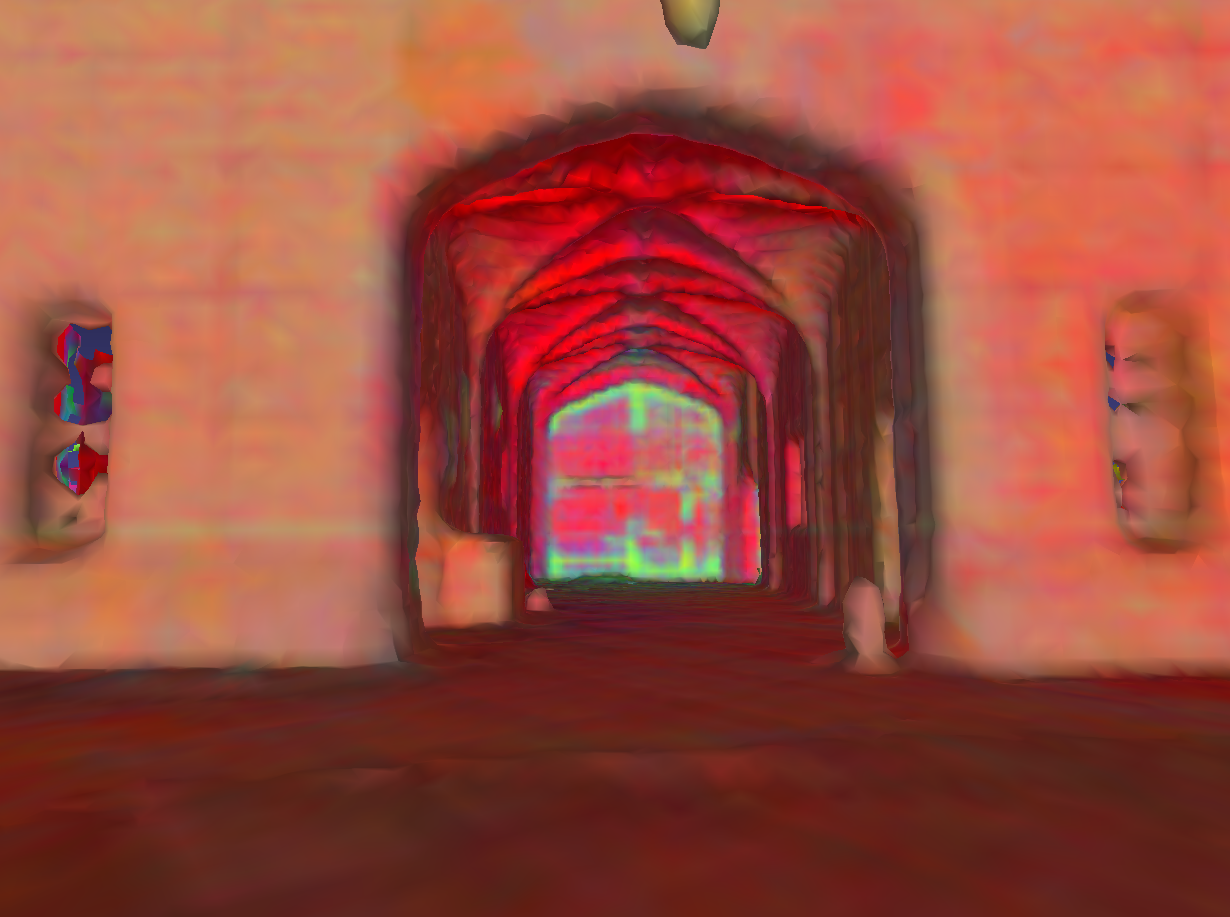}}
  \subfloat[ \label{teaser3}]{\includegraphics[height=3.5cm]{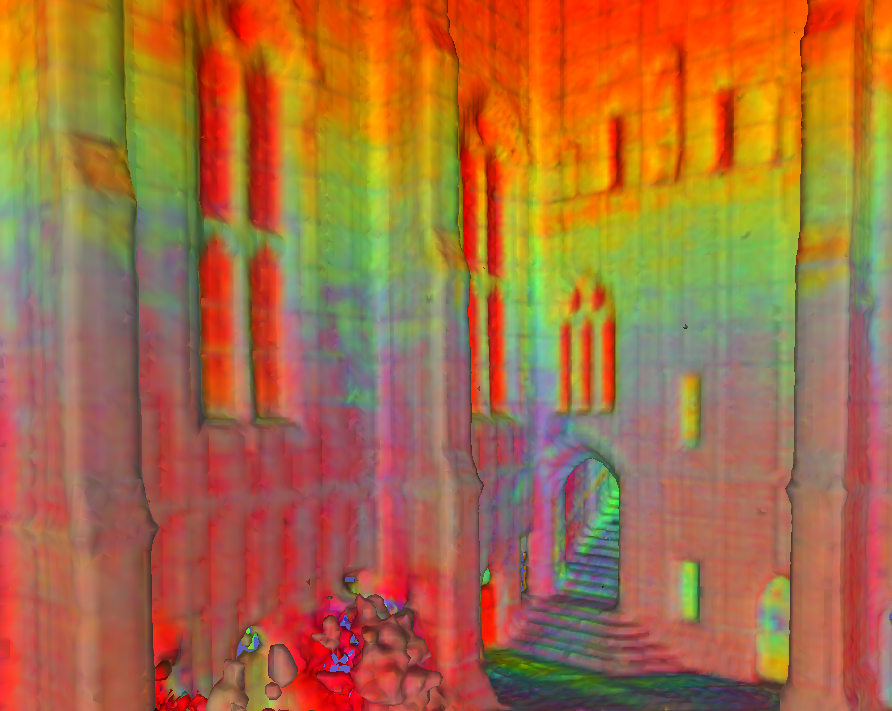}}
  }
  \resizebox{0.8\linewidth}{!}{
  \subfloat{\includegraphics[height=6cm]{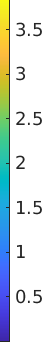}}
  \setcounter{subfigure}{3}
  \subfloat[ \label{teaser4}]{\includegraphics[height=6cm]{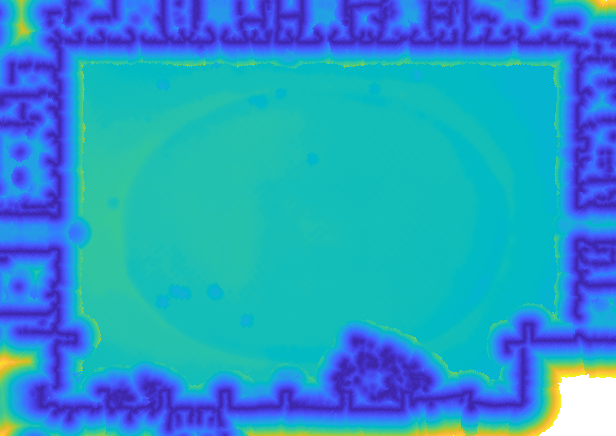}}
  }
  \caption{3D dense reconstruction and distance field from VDB-GPDF framework. a) Shows the incrementally built mesh coloured with the fused LiDAR intensity. Zooming in to visualise b) the ceiling inside the corridor and c) the stairs and windows around the corner of the quad. d) Shows a horizontal slice of the inferred distance field 0.9m above the ground.}
  \label{teaser}
\end{figure}

\IEEEpeerreviewmaketitle

\section{Introduction}\label{sec:introduction}
\IEEEPARstart{R}{obots} understand and interact with the world in a meaningful and efficient manner via effective representations. Building a representation that underpins perception, control, navigation, learning, and manipulation is fundamental but has key requirements that include: 1) high accuracy to reflect the true nature of the unknown environment, 2) efficiency to allow online performance, 3) scalability to handle large-scale scenarios, 4) adaptability to deal with dynamic changes, 5) compatibility with varying sensors, 6) flexibility to provide the needed output space, 7) robustness to ensure reliable operation in the presence of noise and 8) being dense and potentially generative to deal with incomplete and discrete data. 

Recent frameworks based on distance field representations that fulfil some of the above-mentioned requirements have been proposed in the robotics literature~\cite{oleynikova_voxblox_2017,han2019fiesta,vizzo2022vdbfusion,zhu_vdb-edt_2021,bai_vdbblox_2023}. In general, these frameworks are dense and efficient enough to run online, can deal up to a certain point with dynamic objects and are catered to any depth sensor. Most of them incrementally build a projective~\cite{oleynikova_voxblox_2017,vizzo2022vdbfusion} or non-projective~\cite{bai_vdbblox_2023} Truncated Signed Distance Field (TSDF) or occupancy map~\cite{han2019fiesta,zhu_vdb-edt_2021} and some can then recover an approximation of the Euclidean Signed Distance Field (ESDF)~\cite{oleynikova_voxblox_2017,han2019fiesta,bai_vdbblox_2023,zhu_vdb-edt_2021}.
Despite being dense volumetric representations, these frameworks do not rely on generative models thus the representation is maintained at the resolution that was built and only contains information where the data was observed. 

On the other hand, Gaussian Process (GP) framework proposed in our previous work~\cite{wu_faithful_2021,wu_log-gpis-mop_2023,gentil_accurate_2023} are probabilistic generative models that are capable of building accurate Euclidean Distance Field (EDF)s. GP-based frameworks, however, suffer from high computational complexity. To solve this issue and allow online incremental mapping works like~\cite{bhoram_online_2019,gentil_accurate_2023,ali2024interactive} use the Octree data structure. The time of accessing the Octree grows with the number of nodes, thus suffering from scalability issues for large datasets. As a fast-access data structure, OpenVDB~\cite{museth2013vdb} was used recently proposed in VDBFusion~\cite{vizzo2022vdbfusion}, an efficient incremental reconstruction framework based on TSDF, however, does not produce the EDF. 

In this paper, we propose a framework that couples the OpenVDB data structure with GP distance fields. The so-called VDB-GPDF framework builds a scene representation that aims to meet all the key requirements mentioned above. It offers the accuracy and scalability to build a global and incremental scene representation that can cater for room-sized indoor scenes and large-scale outdoor datasets and is compatible with depth sensing such as RGB-D cameras or LiDARs. 
VDB-GPDF deals implicitly with dynamic changes, handles probabilistically noisy measurements, and has the ability to complete data at the required resolution.
Our framework can provide multiple outputs such as accurate Euclidean distance, its gradients, surface properties like colour, intensity or instance segmentation, and a dense mesh with informative textures. The key aspect of this work is the local VDB structure that contains a Local GP Signed Distance Field (L-GPDF), a temporary latent model that enables queries at arbitrary testing points on the current field of view. Followed by a probabilistic fusion that merges the inferred values of the testing points into a global VDB. 
The contributions of this paper are as follows:

\begin{itemize}
\item A seamless coupling of the GPDF with VDB data structure to enable online incremental performance for large-scale mapping.

\item A probabilistic fusion that given the current measurements fits a local GPDF, which allows us to fuse using its mean and uncertainty with the global ESDF and other surface properties.

\item A generative model that efficiently produces the ESDF via a global GPDF trained based on the estimated surface which enables downstream applications to query accurate distance and gradients. Our code is publicly available\footnote{{\tt \url{https://github.com/UTS-RI/VDB_GPDF}}}. 
\end{itemize}

We evaluate our VDB-GPDF against the state-of-the-art frameworks showing comparable reconstruction performance and
superior efficiency and accuracy of the distance field \edit{especially in large-scale datasets}.

\section{Related Work}\label{sec:related work}
Efficient dense volumetric mapping is appealing for the ability to incrementally build a representation of an unseen environment that can be used for localisation, visualisation, navigation and manipulation in particular when the output space is an ESDF. Frameworks like Voxblox~\cite{oleynikova_voxblox_2017} and Voxfield~\cite{pan_voxfield_2022} propose solutions to compute the ESDF as a post-processing step from the projective or non-projective TSDF respectively through wavefront propagation using BFS (Breadth-First Search) algorithm. Using occupancy instead of TSDF, FIESTA~\cite{han2019fiesta} also obtains the ESDF from the occupancy map using multiple customised data structures. 
Employing a less customised data structure, VDB-EDT adopts OpenVDB~\cite{museth2013openvdb,museth2013vdb} for occupancy mapping and a distance transform function to represent the EDF hierarchically~\cite{zhu_vdb-edt_2021}.
Other than occupancy mapping, VDBFusion~\cite{vizzo2022vdbfusion} combines OpenVDB with TSDF fusion to generate volumetric dense maps efficiently. However, VDBFusion does not generate the ESDF as an output. 
VDBblox~\cite{bai_vdbblox_2023} adopts the OpenVDB structure to maintain efficiency. However, VDBblox continues to use TSDF fusion and then propagates it to approximate the ESDF as in~\cite{pan_voxfield_2022}. 

Instead of TSDF or occupancy fusion as all these approaches, we propose to fuse directly on ESDF using an L-GPDF, which allows us to query the Euclidean distance with uncertainty for any point in the field of view. The fusion is then done by merging the inferred EDSF values into the global OpenVDB structure. 
In Table.~\ref{comparison_table}, we summarise the most popular online dense volumetric mapping frameworks and add the proposed framework to highlight its difference and novelty. 
{\small \begin{table}[t]
\caption{Comparison of online dense mapping frameworks}
\begin{center}
\begin{tabular}{|c|c|c|c|c|c|}
\hline
\textbf{Framework} & \textbf{Data Structure} & \textbf{Fusion} & \textsl{Mesh} & \textsl{ESDF} \\
\hline 
\textsl{VDBFusion~\cite{vizzo2022vdbfusion}} & OpenVDB & TSDF & \markgreen{\Checkmark} & \markred{\ding{55}} \\
\hline
\textsl{VDB-EDT~\cite{zhu_vdb-edt_2021}} & OpenVDB & Occupancy & \markred{\ding{55}} & \markgreen{\Checkmark} \\
\hline
\textsl{VDBblox~\cite{bai_vdbblox_2023}} & OpenVDB & TSDF & \markgreen{\Checkmark} & \markgreen{\Checkmark} \\
\hline
\textsl{Voxblox~\cite{oleynikova_voxblox_2017}} & Hash Map & TSDF & \markgreen{\Checkmark} & \markgreen{\Checkmark} \\
\hline
\textsl{Voxfield~\cite{pan_voxfield_2022}} & Hash Map & TSDF & \markgreen{\Checkmark} & \markgreen{\Checkmark} \\
\hline
\textsl{FIESTA~\cite{han2019fiesta}} & Multiple & Occupancy & \markred{\ding{55}} & \markgreen{\Checkmark} \\
\hline
\hline
\textsl{VDB-GPDF (ours)} & OpenVDB & ESDF & \markgreen{\Checkmark} & \markgreen{\Checkmark} \\
\hline
\end{tabular}
\end{center}
\label{comparison_table}
\vspace{-5ex}
\end{table}}
GP-based distance fields have been proposed as a probabilistic and generative model to represent complex environments~\cite{bhoram_online_2019,wu_faithful_2021}. 
In our prior work~\cite{wu_faithful_2021,wu_log-gpis-mop_2023}, we proposed a theoretical sound formulation to estimate the Euclidean distance field by applying the logarithmic transformation to a GP formulation. 
To improve the accuracy of the distance estimate, we later introduced the reverting GP distance field~\cite{gentil_accurate_2023}. 
Although GP-based frameworks are computationally intensive, Octree structures have enabled fast computations in~\cite{bhoram_online_2019,gentil_accurate_2023} and more recently in~\cite{ali2024interactive} for online incremental mapping. However, Octrees expand the nodes over time and require tree traversal, resulting in a less scalable data structure for large scenarios. Recently, methods like the Gaussian mixture model~\cite{goel2024distance} have been introduced to model an EDF using fused point clouds or batch mapping, but a substantial gap remains in achieving incremental performance. Via the VDB data structure, our proposed approach facilitates efficiency and scalability while maintaining the advantages of the GPDFs.

Beyond GPs, other generative methods for dense volumetric mapping have recently been proposed in the literature. A large number of works have examined the potential of deep learning techniques for EDF representations. Inspired by~\cite{park_deepsdf_2019}, the work in~\cite{gropp_learning_shapes_2020} proposes using an implicit neural representation with Eikonal regularisation to approximate the ESDF for points distant from the surface. The work in~\cite{pantic_NeRF_planning_2022} introduced learning an EDF approximation from Neural Radiance Fields (NeRFs) using occupancy. Similarly, inspired from~\cite{gropp_learning_shapes_2020}, iSDF~\cite{ortiz_isdf_2022} is proposed to use a neural signed distance field for mapping.
To achieve efficient performance in the learning approach, DI-Fusion~\cite{huang_di_fusion_2021} proposes an online incremental SDF using clustering blocks, and HIO-SDF~\cite{vasilopoulos_hio_sdf_2023} applies hierarchical data structure. 
Despite the advantages of this type of continuous representation with low memory consumption, all these approaches need extensive pre-training to enable incremental mapping and rely on high-performance GPU capabilities.

\section{Preliminaries}\label{sec:background} 
\subsection{Gaussian Process Distance Field}
\label{GP-distance-field}
GP~\cite{GPbook} is a non-parametric regression approach that models a distribution over functions. We use the so-called GP Distance Field method originally presented in \cite{gentil_accurate_2023} to model our distance field.
Consider a surface $\mathcal{S}$ in a Euclidean space $\mathbb{R}^D$, and a set of discrete observations of $S$ as $\mathbf{y}=\left\{y_{j} \right\}_{j=1}^{J} \in \mathbb{R}$ taken at locations $\mathbf{X}=\{\abscissavec_j\}_{j=1}^{J} \in \mathbb{R}^{D}$.
By modelling the latent scalar field which can be interpreted as an occupancy field $o(\abscissavec): \mathbb{R}^D \mapsto \mathbb{R}$ of the space with a GP as $o \sim \mathcal{GP}\left(0, k(\abscissavec, \abscissavec')\right)$, it is possible to infer $\hat{\occ}(\abscissavec_*)$ at any location in the space.
Let us arbitrarily define the occupied area to be equal to 1. Therefore, $\mathbf{y}$ is equal to $\mathbf{1}$ for the GP latent inference
\begin{align}
    \hat{\occ}(\abscissavec_*)& = \kernelvecraw{\abscissavec_*\abscissamat} \left(\kernelmatraw{\abscissamat\abscissamat} + \sigma_{\occ}^2\identitymat \right)^{-1} \mathbf{1}.
    \label{eq:latent_inference}
\end{align}
where $\sigma_{\occ}^2$ is the variance of noise.
The uncertainty is then inferred using
\begin{align}
    \hat{u}(\abscissavec_*)& = \kernelvecraw{\abscissavec_*\abscissavec_*}+\kernelvecraw{\abscissavec_*\abscissamat} \left(\kernelmatraw{\abscissamat\abscissamat} + \sigma_{\occ}^2\identitymat \right)^{-1} \kernelvecraw{\abscissamat\abscissavec_*}.
    \label{eq:latent_variancee}
\end{align}

The distance field $\distfield(\abscissavec_*)$ given any location $\abscissavec_*$ is obtained by applying the \emph{reverting} function $\revert$ to the latent field as $\distfield(\abscissavec_*) = \revert\left(\hat{\occ}\left(\abscissavec_*\right)\right)$.
Considering the unscaled square exponential kernel $\kernel{}{\abscissavec}{\abscissavec'} = \exp\left(-\frac{\lVert \abscissavec - \abscissavec' \rVert{}^2}{2\lengthscale^2}\right)$, with the kernel's length scale $\lengthscale$, the reverting function is defined as 
\begin{align}
    \distfield(\abscissavec_*) = \sqrt{-2\lengthscale^2\log\left(\hat{\occ}\left(\abscissavec_*\right)\right)}.
    \label{eq:dist_field_final}
\end{align}
As explained in \cite{gentil_accurate_2023}, the resulting distance field is not a GP due to the non-linear reverting operation.
Thus its variance is not readily available throughout $\mathbb{R}^3$. By simply propagating the latent field's uncertainty through the first-order Taylor expansion of the reverting function as
\begin{equation}
    \hat{v}(\abscissavec_*)=\Lambda \hat{u}(\abscissavec_*) \Lambda^{\top}, 
    \text{with}\ \Lambda = \frac{-\lengthscale}{\hat{\occ}(\abscissavec_*)\sqrt{-2\log\left(\hat{\occ}(\abscissavec_*)\right)}},
    \label{eq:variance_final}
\end{equation}
the distance uncertainty away from the surface grows indefinitely.
However, close to the surface, as required here for the fusion process, $\hat{v}(\abscissavec_*)$ provides a measurement of the uncertainty\footnote{As stated in \cite{gentil_accurate_2023}, the latent field is not bounded between 0 and 1. The value of $\hat{\occ}(\abscissavec_*)$ is capped to prevent any singularity on $\distfield(\abscissavec_*)$ and $\hat{v}(\abscissavec_*)$.}.

To estimate the gradient of the distance field, we simply apply the differentiation (linear) operator as
\begin{equation}
\begin{aligned}
& \nabla\hat{d}\left(\abscissavec_*\right)=\nabla\mathbf{k}_{\mathbf{x_*} \mathbf{X}}\left(\mathbf{K}_{\mathbf{X} \mathbf{X}}+\sigma_{\occ}^{2} \mathbf{I}\right)^{-1} \mathbf{1}, 
\end{aligned}
\label{eq:linear_operator_mean}
\end{equation}
where $\nabla\mathbf{k}_{\mathbf{x_*} \mathbf{X}}$ is the partial derivative of the kernel matrix with respect to $\mathbf{x}$.

\subsection{OpenVDB Data Structure}\label{sec.openvdb}
\edit{OpenVDB (Volumetric, Dynamic grid that shares several characteristics with B+trees) is a hybrid data structure. It splits the space into hash tables and stores a hierarchical tree inside each table, resulting in sharing the advantages of the hashing method.} OpenVDB~\cite{museth2013vdb,museth2013openvdb} is used to enable efficient, large-scale, dense incremental mapping. This hierarchical tree-like structure allows us to efficiently divide GP into clusters with constant computational complexity to access them. 

The B-tree-like three-dimensional VDB data structure comprises sparse collections of voxel blocks in four layers: root nodes, leaf nodes, and usually two levels of internal nodes. The default configuration is typical $2^3\times2^3\times2^3$ voxels in each leaf node, $2^4\times2^4\times2^4$ of leaf nodes in a first-layer of internal node, and $2^5\times2^5\times2^5$ of the first-layer of internal nodes in a second-layer of internal node, Therefore, a second-layer of internal node subsumes a three-dimensional block of voxels of size $4096\times4096\times4096$. 
Unlike the internal nodes and leaf nodes, the number of children of the root node is not explicitly fixed. As default, each root node has $4$ second-layer internal nodes. This hierarchical tree structure is fixed with four height layers across the involved tree, maintaining a shallow and broad representation compared to octrees. The constant height of the VDB tree allows constant and fast traversing from the root node to the leaf node. Any random access can operate at a consistent computational complexity in average $\mathcal{O}(1)$. Furthermore, VDB substantially reduces contemporary CPU memory consumption by hierarchically allocating the nodes even with extreme grid resolution. 

The memory and time efficiency properties of VDB~\cite{museth2013openvdb} are highly suitable for our online mapping frameworks.
We treat each leaf node in VDB as a GP cluster, and a leaf iterator is used to traverse voxels and train GP clusters separately.
\begin{figure*}[t]
	\centering
	\resizebox{1.0\linewidth}{!}{
	\includegraphics[]{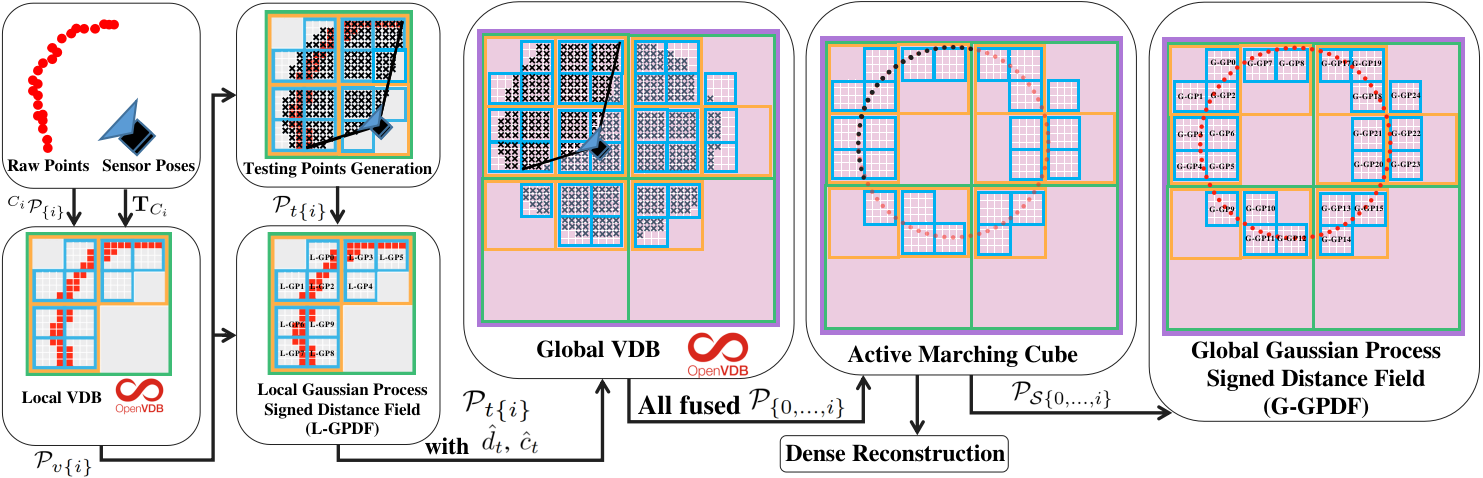}
	}
	\caption{Block Diagram of the proposed VDB-GPDF framework. We first model the temporary Local Gaussian Process Signed Distance Field (L-GPDF) and surface properties using $\mathcal{P}_{v\{i\}}$. A set of testing points along the ray from the sensor origin to $\mathcal{P}_{v\{i\}}$ are generated to query the distance and surface properties inferences of L-GPDF. 
    Each point in $\mathcal{P}_{t\{i\}}$ with its inferred $\hat{d}_{t}$ and $\hat{c}_{t}$ are fused with the full map represented by a global VDB. Then we have all fused voxels $\mathcal{P}_{\{0,...,i\}}$. The marching cube is applied on the active voxels in $\mathcal{P}_{\{0,...,i\}}$ to update the zero crossing points $\mathcal{P}_{\mathcal{S}\{0,...,i\}}$ and dense reconstruction. The global Gaussian Process Signed Distance Field (G-GPDF) is modelled by $\mathcal{P}_{\mathcal{S}\{0,...,i\}}$ in each leaf node in the global VDB separately. 
    }
    \vspace{-2ex}
	\label{fig:overview}
\end{figure*}
\section{VDB-GPDF Framework}\label{sec:method}
To incrementally build and maintain a persistent, efficient, large-scale, adaptable, robust and dense distance field map of the environment with depth sensors, we propose to couple GPDF~\cite{gentil_accurate_2023} with local and global VDB data structures. Fig.~\ref{fig:overview} shows the diagram of our proposed VDB-GPDF framework. 

Depth sensor data is captured as a raw point cloud ${ }^{{C}_{{i}}} \mathcal{P}_{\{i\}}$ in sensor frame ${C}_{{i}}$ at current time $i$. The estimated pose is used to project the point cloud from current sensor frame ${{C}_{{i}}}$ to world frame ${W}$ given the homogeneous transformation matrix $\mathbf{T}_{{C}_{{i}}}$ yielding a point cloud $\mathcal{P}_{\{i\}}$ in the world reference frame. The raw points $\mathcal{P}_{\{i\}}$ are voxelised into the voxels denoted as $\mathcal{P}_{v\{i\}}$ via a local hierarchical VDB. Thus, $\mathcal{P}_{v\{i\}}$ is divided into local clusters maintained by the VDB leaf nodes. For each leaf node, the centres of voxels are used as training points to model GPs separately for distance and surface properties. All GPs together form the temporary L-GPDF along with surface properties. 
In addition, $\mathcal{P}_{v\{i\}}$ is passed to the testing points generation block. A ray-cast operation from the current sensor origin to each voxel in $\mathcal{P}_{v\{i\}}$ is performed to generate a set of voxels. These voxels are then used as the testing points $\mathcal{P}_{t\{i\}}$ to query the L-GPDF. Let us denote each point in $\mathcal{P}_{t\{i\}}$ as $\mathbf{x_*}$. Note that the distance sign of each $\mathbf{x_*}$ is computed given the sensor origin and $\mathcal{P}_{v\{i\}}$.
For each querying point $\mathbf{x_*}$, the closest node in L-GPDF is found to compute the GP distance inference $\distfield_t(\mathbf{x_*})$, surface properties $\hat{c}_t(\mathbf{x_*})$ and uncertainty.

The $\hat{d}_{t}$ and $\hat{c}_t$ estimates (mean and uncertainty) at each location in $\mathcal{P}_{t\{i\}}$ are then fused with the existing map maintained by a global VDB grid, which has all fused voxels $\mathcal{P}_{\{0,...,i\}}$. 
The leaf nodes in the global VDB grid that participated in the fusion procedure are marked as active. Marching cubes algorithm~\cite{marching} is then performed for the voxels in the active nodes to recover the dense reconstruction. The zero crossing points $\mathcal{P}_{\mathcal{S}\{0,...,i\}}$ in active nodes and previous active nodes are used to generate the G-GPDF. Similar to L-GPDF, each leaf node is trained separately. The G-GPDF is available to be queried for any given points $\mathcal{P}_{o}$ for downstream applications that require accurate distance $\hat{d}_{o}$ and gradients $\nabla \hat{d}_{o}$, \emph{e.g.}, navigation, path planning, localisation and manipulation.

\subsection{Local Gaussian Process Signed Distance Field}
We propose to efficiently voxelise the raw points $\mathcal{P}_{\{i\}}$ in the world frame into a set of voxels via a local hierarchical VDB. The method groups all raw points projecting into the same voxel. Therefore, a dense raw point cloud $\mathcal{P}_{\{i\}}$ is represented by the centres of voxels, denoting as $\mathcal{P}_{v\{i\}}$. We simply perform the further fusion and integration once for each voxel. Taking advantage of efficient access to VDB, the voxelisation procedure is fast and $\mathcal{P}_{v\{i\}}$ is automatically divided into leaf nodes maintained by the VDB.  

We iterate each leaf node of the local VDB for voxels, which are used as training points to formulate each node as a GP separately. All GPs in local VDB consist of the temporary L-GPDF along with surface properties. As explained in Section.~\ref{GP-distance-field}, given locations of voxels, we compute the distance field inference and variance via Eq.~\ref{eq:dist_field_final} and Eq.~\ref{eq:variance_final}. In addition, given the voxels' surface properties of each leaf node, we propose to model the surface properties inference:
\begin{align}
    \hat{c}_t(\abscissavec_*)& = \kernelvecraw{\abscissavec_*\abscissamat} \left(\kernelmatraw{\abscissamat\abscissamat} + \sigma_{c}^2\identitymat \right)^{-1} \mathbf{c}.
    \label{eq:r_inference}
\end{align}
Therefore, L-GPDF is ready for the testing points to query the distance, surface properties and variance. Note that the temporary and local L-GPDF is re-modelled at every frame $i$. 
\subsection{Testing Points Generation}\label{Testing Points Generation}
Testing points generation method takes $\mathcal{P}_{v\{i\}}$ and sensor origin as input. Here, we adopt a similar idea as the so-called free space carving method~\cite{oleynikova_voxblox_2017,vizzo2022vdbfusion} to update the distance information for the voxels within the current sensor field of view. 
Free-space carving has the ability to update the dynamically changing objects in the scene. \edit{We update voxels with previously and newly observed surfaces.} We propose three main improvements to enhance efficiency while achieving comparable reconstruction results to the conventional free-space carving used in most volumetric mapping methods. 

First, we propose a voxelisation method to ray-cast from the current sensor origin to each voxel in $\mathcal{P}_{v\{i\}}$. This results in a faster traverse than ray-casting for each raw point cloud $\mathcal{P}_{\{i\}}$ and produces comparable zero crossing. Secondly, the naive ray-casting usually iterates and updates all voxels intersected by the ray in order. This may lead to updating the voxel without any surface measurement or the same voxel multiple times due to having different ray intersections. In contrast, we only query the voxel, which has previous surface measurements in the field of view. In this way, the existence of the previous surface is updated. Therefore, the dynamic objects are handled implicitly and efficiently.
Note that testing all points in the frustum is time-consuming and not necessary in our framework. Despite we can get the fused ESDF this does not produce a generative model that can be queried anywhere in the space. Therefore, we choose to infer only the relevant testing points for fusion, making this process more efficient and later generating the accurate ESDF using G-GPDF as explained in the next section.
Thirdly, we compute the surface normals given $\mathcal{P}_{v\{i\}}$ and generate voxels along the normal direction up to a certain distance even outside the frustum. When the sensor ray is close to being parallel to the observed surface, the naive ray-casting cannot generate enough testing voxels around the surface. Having extra voxels along the normal direction yields a complete reconstruction.

We use $\mathcal{P}_{t\{i\}}$ to query the distance through the L-GPDF. 
For each $\mathbf{x_*}$ in $\mathcal{P}_{t\{i\}}$, we find the closest node in L-GPDF and use the node GP to compute the GP distance inference $\hat{d}_{t}(\mathbf{x_*})$, surface property inference $\hat{c}_{t}(\mathbf{x_*})$, and variances $\hat{v}_{t}(\mathbf{x_*})$ and $\hat{w}_{t}(\mathbf{x_*})$. 
Thanks to the available Euclidean distance inference of L-GPDF, we only query each voxel once to perform the fusion later. This leads to a comparable result of the final reconstruction but more efficiently than TSDF fusion. Note that we also compute the distance sign given the sensor origin.

\subsection{Data Fusion}
After querying the testing points from the L-GPDF, we have the signed distance inference $\hat{d}_{t}(\mathbf{x_*})$, $\hat{c}_{t}(\mathbf{x_*})$ and variance $\hat{v}_{t}(\mathbf{x_*})$ $\hat{w}_{t}(\mathbf{x_*})$ for each $\mathbf{x_*}$ in $\mathcal{P}_{t\{i\}}$. 
Unlike the commonly used TSDF fusion that uses constant or drop-off weights to fuse the projective (or non-projective) distance, we directly query our probabilistic L-GPDF to fuse the distance and surface properties through its mean and variance in the global VDB. Note that the fusion is mostly done around the surface, as discussed in Section~\ref{sec:background}.

The global VDB efficiently accesses the previous distance mean and variance at $\mathbf{x_*} \in\mathcal{P}_{t\{i\}}$, which are denoted as $\boldsymbol{D}_{i-1}(\mathbf{x_*})$ and $\boldsymbol{V}_{i-1}(\mathbf{x_*})$ respectively. Then new $\boldsymbol{D}_{i}(\mathbf{x_*})$ and $\boldsymbol{V}_{i}(\mathbf{x_*})$ are estimated given the queried $\hat{d}_{t}(\mathbf{x_*})$ $\hat{v}_t(\mathbf{x_*})$ of L-GPDF.
\markred{We then perform fusion by using the standard weighting average,
\begin{equation}
\boldsymbol{D}_i(\mathbf{x_*}) = \alpha \cdot \hat{d}_{t}(\mathbf{x_*}) + \beta \cdot \boldsymbol{D}_{i-1}(\mathbf{x_*}), 
    \ \boldsymbol{V}_i(\mathbf{x_*}) = \gamma.
\end{equation}
where the probabilistic weights are given by,
\begin{equation}
\begin{aligned}
\alpha &=\frac{\boldsymbol{V}_{i-1}(\mathbf{x_*})}{\boldsymbol{V}_{i-1}(\mathbf{x_*})+\hat{v}_t(\mathbf{x_*})}, \ \beta =\frac{\hat{v}_t(\mathbf{x_*})}{\boldsymbol{V}_{i-1}(\mathbf{x_*})+\hat{v}_t(\mathbf{x_*})}, \\ \gamma &=(\boldsymbol{V}_{i-1}(\mathbf{x_*})^{-1}+\hat{v}_t(\mathbf{x_*})^{-1})^{-1}.
\end{aligned}
\label{eq: probabilistic fusion}
\end{equation}
Alternatively, we can also use our variance to update the weights following~\cite{tsdf}, 
\begin{equation}
\begin{aligned}
\alpha &=\frac{1-\hat{v}_t(\mathbf{x_*})}{\boldsymbol{V}_{i-1}(\mathbf{x_*})+1-\hat{v}_t(\mathbf{x_*})}, \ \beta = \frac{\boldsymbol{V}_{i-1}(\mathbf{x_*})}{\boldsymbol{V}_{i-1}(\mathbf{x_*})+1-\hat{v}_t(\mathbf{x_*})}, \\ \gamma &=\boldsymbol{V}_{i-1}(\mathbf{x_*})+1-\hat{v}_t(\mathbf{x_*}).
\end{aligned}
\label{eq: tsdf_fusion}
\end{equation}
Similarly, the surface properties are fused using the same expressions as above but with their respective variance $\hat{w}_{t}(\mathbf{x_*})$.}
\subsection{Global Gaussian Process Signed Distance Field}
After the fusion process, the leaf nodes in the global VDB grid are marked as active nodes. We perform the marching cubes for the voxels of active nodes and for computing the dense reconstruction with surface properties. The zero crossing points in current active nodes $\mathcal{P}_{\mathcal{S}\{i\}}$ then merge into $\mathcal{P}_{\mathcal{S}\{0,...,i\}}$. Zero crossing points in each node are used to train separate GPs. The combination of all GPs forms G-GPDF.

Note that when querying the G-GPDF, we only search for a certain number of closest nodes through a kd-tree for each querying point and only perform the corresponding GP training and inference for these nodes. In this way, we avoid training unnecessary GP nodes which no points are querying them. This will avoid redundancy in computing the GP of the entire field. In addition, we use the smooth minimum function for the distance field to ensure accurate global consistency. 

When partitioning the full map into leaf nodes in the VDB data structure, modelling the clustering GP of each node suffers from discontinuity around the boundaries. A common solution is to have overlapping parts with neighbouring nodes. However, this method extends the volume of each node therefore the GP training becomes more complex in terms of computations. Instead, we address the problem by using the smooth minimum function to estimate the continuous distance field. We search for the number of $Q$ clustered GP nodes in the map and recover the distance as,
\begin{equation}
\distfield(\abscissavec_*) = \frac{\sum_{q=1}^Q\distfield_q(\abscissavec_*) \exp \left(\lambda\distfield_q(\abscissavec_*)\right)}{\sum_{q=1}^Q \exp \left(\lambda\distfield_q(\abscissavec_*)\right)},
\end{equation}
where $\lambda$ is a large positive value to control the evenness of the smooth minimum. We use $100$ for $\lambda$ in this paper. $\distfield_q(\abscissavec_*)$ is the distance inference at querying location $\abscissavec_*$ given the located GP node with the index of $q$. For the gradient, we directly average the inferred gradients from the $Q$ GP nodes.

\section{Evaluation}\label{sec:evaluation}
To evaluate the proposed VDB-GPDF framework, we quantitatively compare the performance for a) fusion output, b) efficiency, c) reconstruction accuracy and d) distance accuracy. Our framework is implemented in C++ based on ROS1 with CPU only. All experiments were run on 12th Gen Intel® Core™ i5-1245U. We use frames 220-820 of the Cow and Lady dataset~\cite{oleynikova_voxblox_2017}, which covers major information of the scene. For the real-world LiDAR dataset, we choose the quad of the Newer College dataset~\cite{ramezani2020newer}. The estimated poses are calculated via the alignment of each frame to the ground truth map. The size of the main quad is roughly $56\times40\times20$ m, with approximately 2000 frames. \edit{We also include the larger simulated LiDAR dataset Mai City~\cite{vizzo2021poisson} sized $700\times120\times3$ m}.
\edit{All the above datasets} have the ground truth map that allows us to perform quantitative evaluations for reconstruction and distance accuracy.
\edit{In addition, we qualitatively validate our framework on the KITTI odometry 07 dataset~\cite{behley2019semantickitti} with dynamic objects in a large-scale map.} 
Our strategy for length scale selection depends on the number of voxels in each leaf node, which in our case is fixed to $2^3\times2^3\times2^3$ voxels. A length scale that correlates two to four voxels is a good trade-off between smoothing and interpolation for our kernel function.

\subsection{Fusion Performance}
\begin{figure}[ht]
  \centering
  \resizebox{0.6\linewidth}{!}{
  \subfloat[TSDF fusion\label{tsdf_fusion}]{\includegraphics[height=4.0cm]{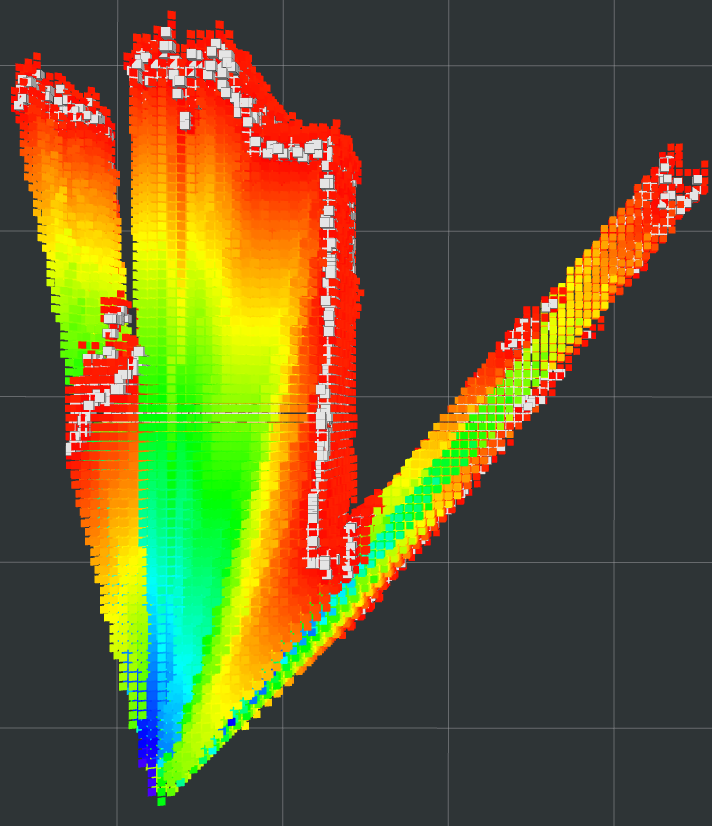}}
  \subfloat[ESDF Fusion \label{esdf_fusion}]{\includegraphics[height=4.0cm]{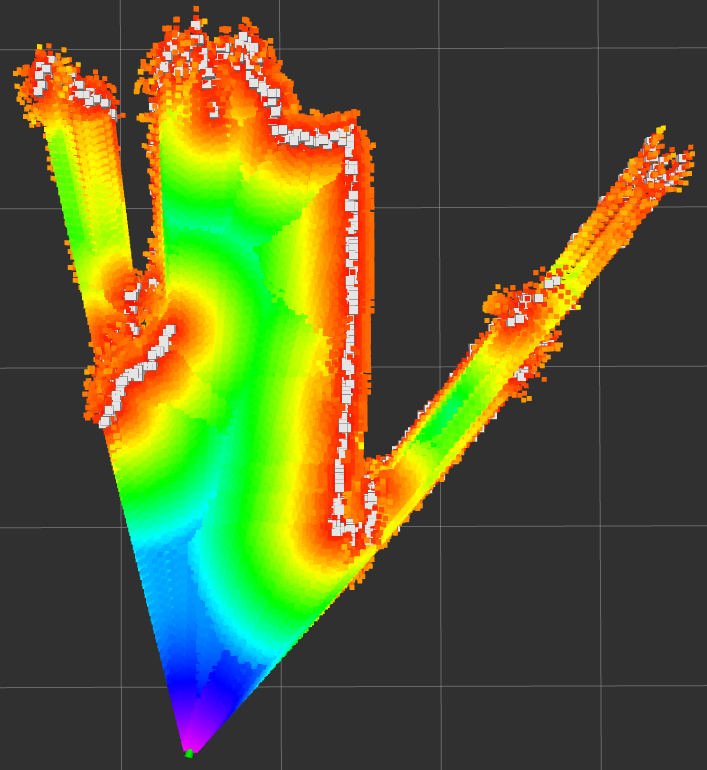}}
  }
  \centering
  \resizebox{0.9\linewidth}{!}{
  \subfloat[Mesh after fusion VDBFusion \label{tsdf_fusion_mesh}]{\includegraphics[height=3.0cm]{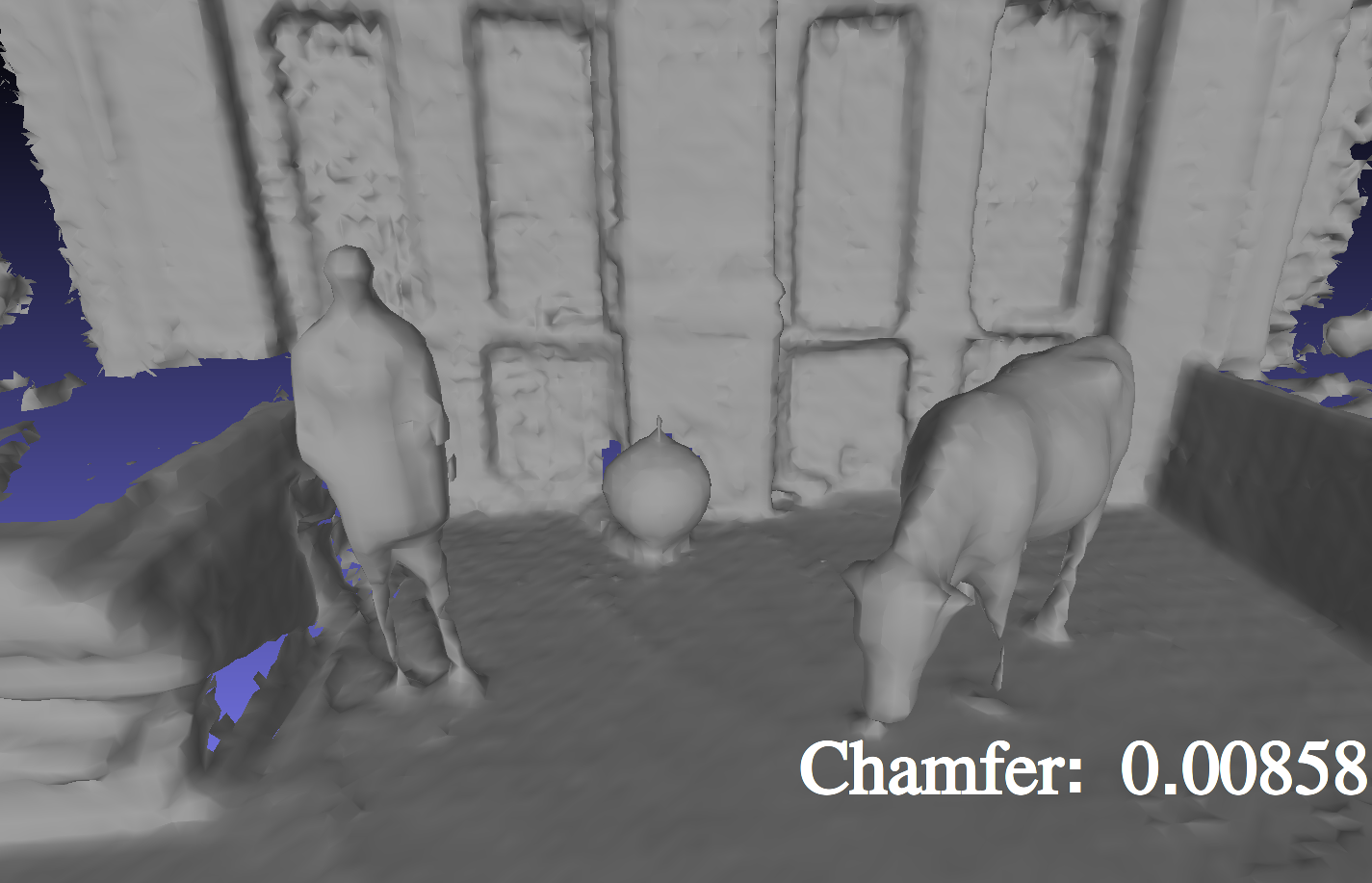}}
  \subfloat[Mesh after fusion VDB-GPDF\label{esdf_fusion_mesh}]{\includegraphics[height=3.0cm]{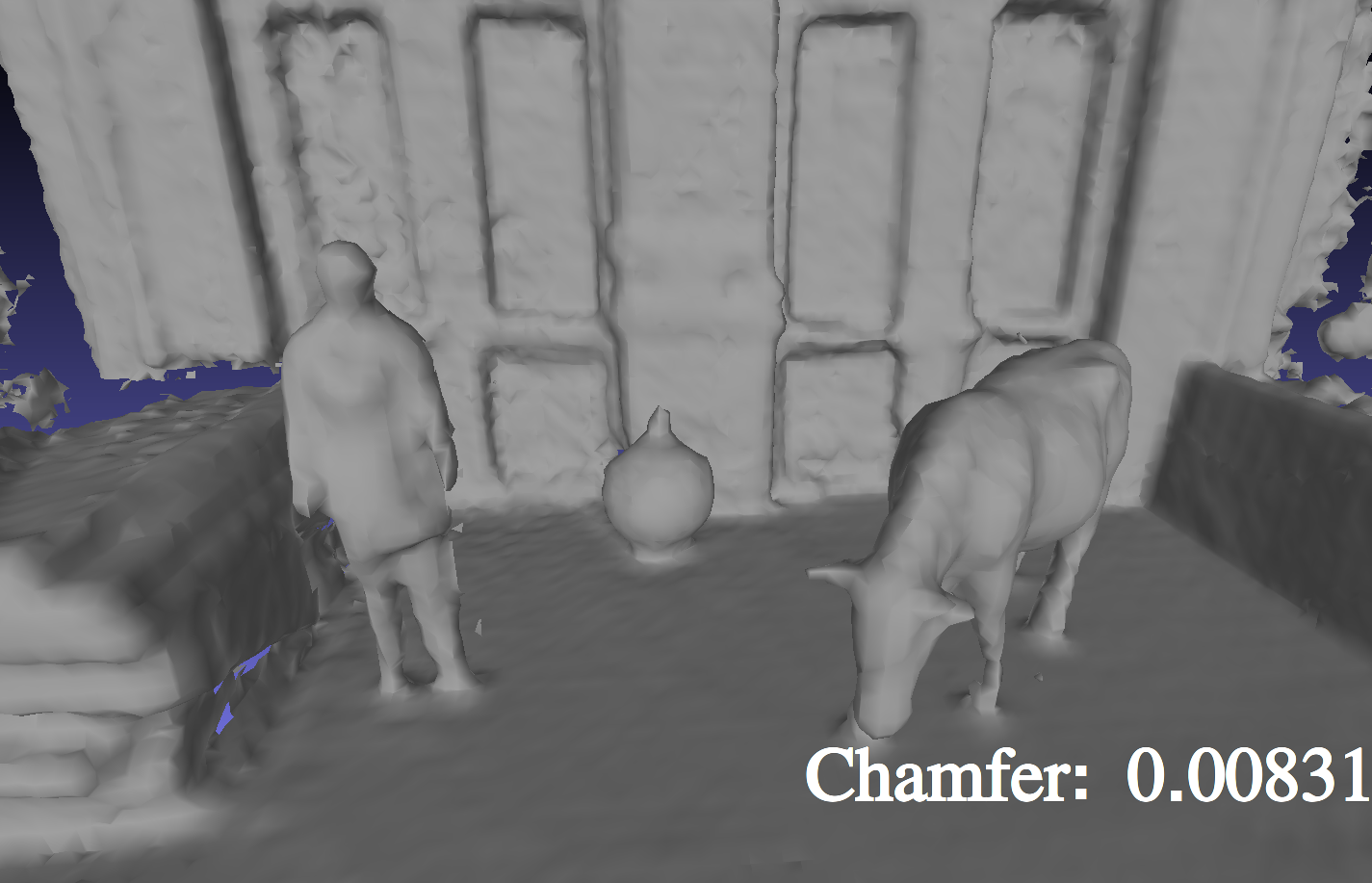}}
  }
  \caption{Qualitative and quantitative comparison of a) the TSDF fusion from VDBFusion and b) our proposed ESDF fusion with full testing points in the frustum. b) shows that our approach reasons directly in the Euclidean space. After the fusion, our reconstruction in d) shows less noise in the wall and is more complete on the mattress and cow legs than c).}
  \label{fusion_figure}
  \vspace{-3ex}
\end{figure}

First, we want to evaluate the proposed fusion. We compare the proposed ESDF fusion with TSDF fusion in VDBFusion.
For the sake of fairness, we apply the same voxelisation method in Sec.~\ref{Testing Points Generation} for VDBFusion as in our proposed method. A qualitative comparison of the TSDF fusion of VDBFusion and the proposed ESDF fusion with full testing voxels in the frustum is demonstrated in Fig.~\ref{tsdf_fusion} and~\ref{esdf_fusion}. The white boxes are the measurements, and the coloured flat squares are the inferred distance values of testing points for fusion in the field of view. The results show that our ESDF fusion produces accurate and natural inferred distance of each point in the frustum. After performing the fusion for all frames, we compare the reconstructed mesh for TSDF fusion and our fusion. Our reconstruction in~\ref{esdf_fusion_mesh} shows less noise in the wall, more complete on the mattress and cow legs than~\ref{tsdf_fusion_mesh}, lower reconstruction error in Chamfer distance.

\subsection{Efficiency}\label{sec:efficiency}
\begin{figure}[ht]
  \centering
  \resizebox{0.9\linewidth}{!}{
  \subfloat{\includegraphics[height=3.0cm]{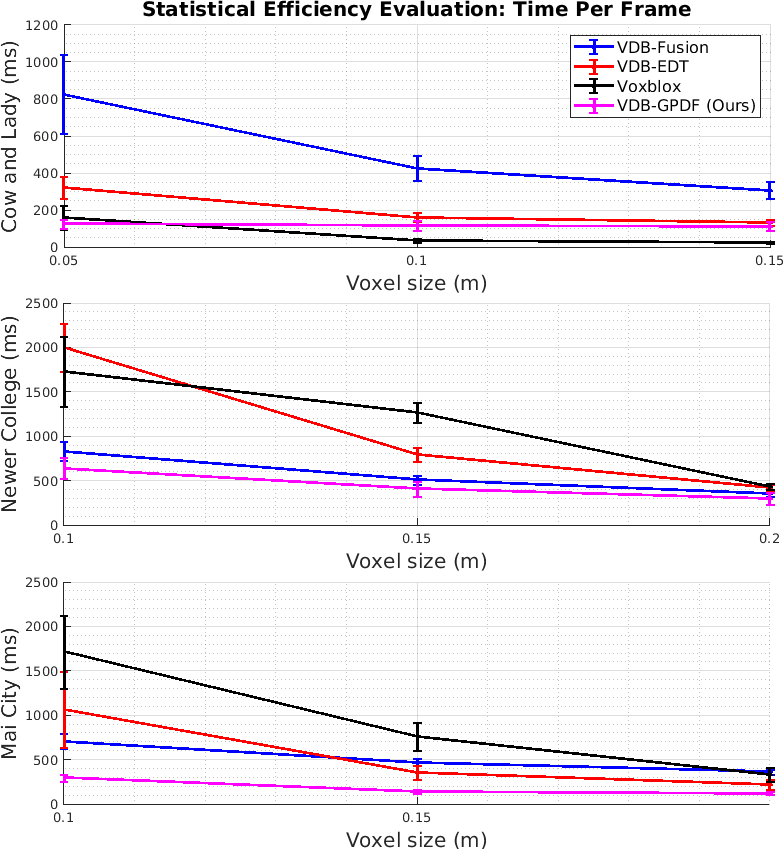}}
  }
  \caption{\edit{Efficiency statistical evaluation of the time per frame for the Cow and Lady (top), the Newer College (middle) and Mai City (bottom) datasets respectively.}}
  \label{time_std}
\end{figure}
We first evaluate the VDB-GPDF efficiency against the state-of-the-art mapping frameworks VDB-Fusion and VDB-EDT, both developed based on OpenVDB data structure\edit{, and Voxblox based on Hash maps. Fig.~\ref{time_std} shows the statistical evaluation of the computational time (mean and standard deviation) for multiple datasets with varying voxel sizes.} Note that the Newer College quad has pedestrians walking by as dynamic objects. All frameworks have free-space carving methods enabled to update the moving objects and noise in the scene. 
For the sake of simplicity, we compute the time for all the integration processes, including fusion and ESDF. As for our framework, we include L-GPDF, incremental fusion and G-GPDF. \edit{For the Cow and Lady, Voxblox is overall more efficient than others while our VDB-GPDF outperforms Voxblox in high resolution. For LiDAR cases, our VDB-GPDF is more efficient than other frameworks with varying resolutions due to our voxelisation method and fewer testing points in the frustum area.}
\begin{figure}[t]
  \centering
  \resizebox{1.0\linewidth}{!}{
  \subfloat[ VDBFusion\label{cow_mesh_vdbfusion}]{\includegraphics[height=2.5cm]{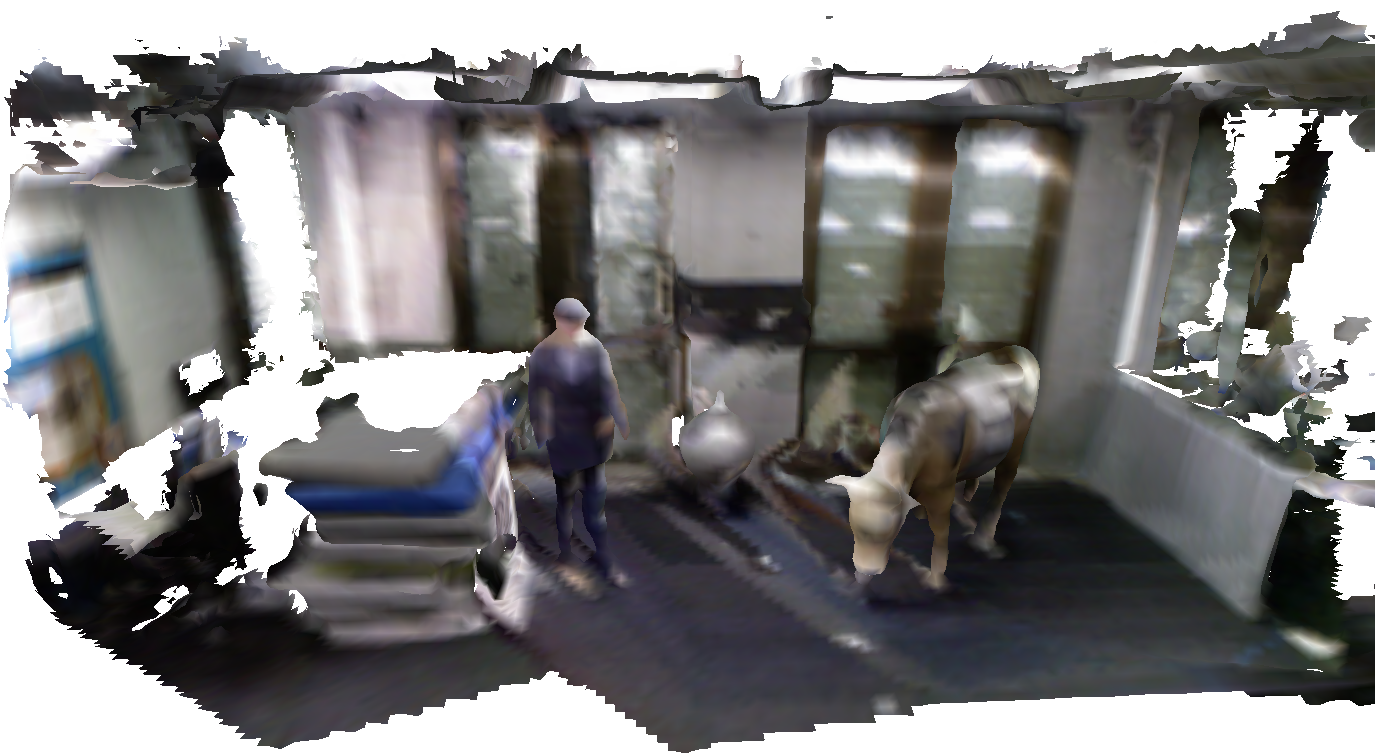}}
  \subfloat[VDB-GPDF\label{cow_mesh_ours}]{\includegraphics[height=2.5cm]{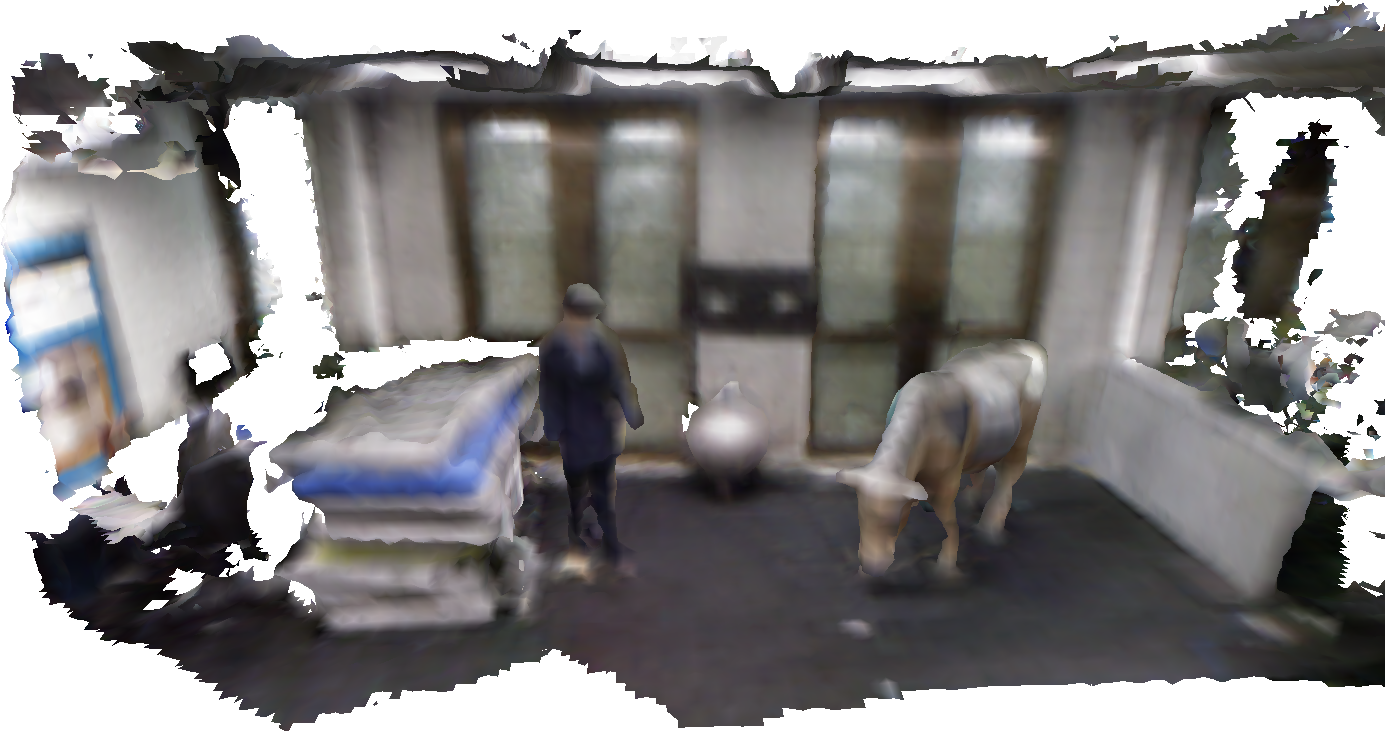}}
  }
  \caption{Qualitative reconstruction comparison of a) VDBFusion and b) our proposed method shows that VDB-GPDF produces a more complete and natural reconstruction, especially when the sensor is in parallel with the observed surface. Our mesh covers on the top of the mattress, and the colour is properly fused for visualisation.}
  \label{cow_mesh_figure}
  \vspace{-3ex}
\end{figure}
\begin{figure}[ht]
  \centering
  \resizebox{\linewidth}{!}{
  \subfloat[ \label{cow:chamfer}]{\includegraphics[height=5.0cm]{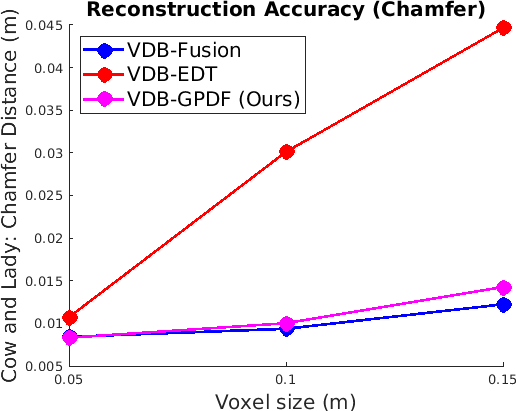}}
  \subfloat[ \label{cow:distance}]{\includegraphics[height=5.0cm]{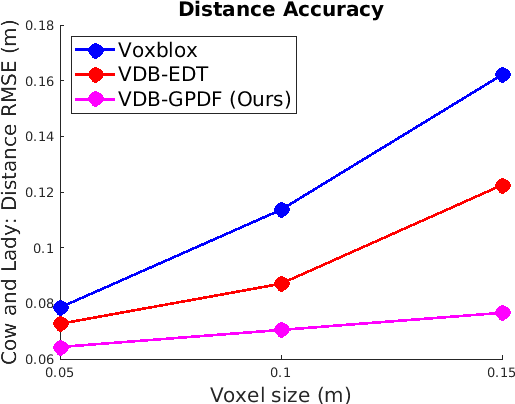}}
  }
  \resizebox{\linewidth}{!}{
  \subfloat[ \label{newer:chamfer}]{\includegraphics[height=5.0cm]{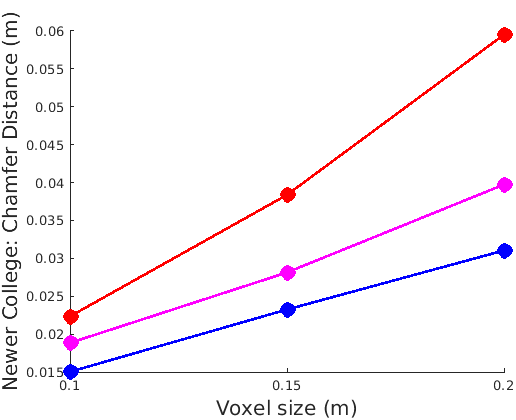}}
  \subfloat[ \label{newer:distance}]{\includegraphics[height=5.0cm]{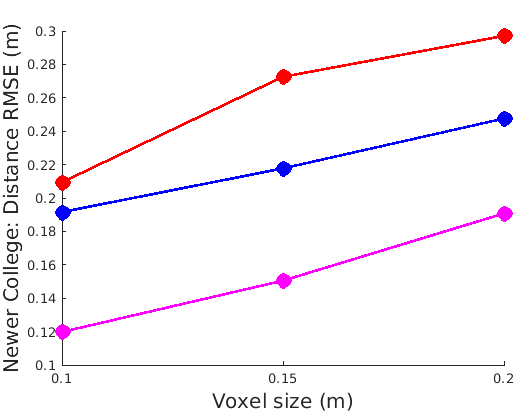}}
  }
  \resizebox{\linewidth}{!}{
  \subfloat[ \label{mai:chamfer}]{\includegraphics[height=5.0cm]{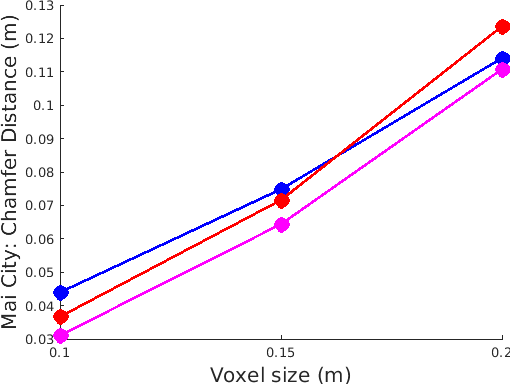}}
  \subfloat[ \label{mai:distance}]{\includegraphics[height=5.0cm]{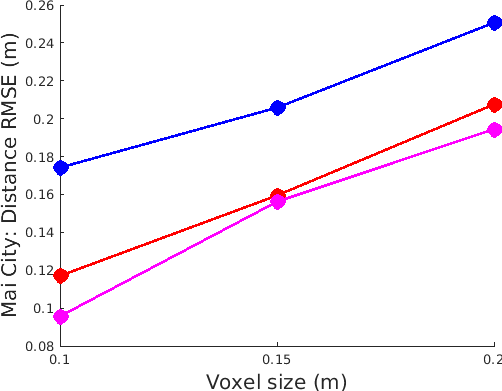}}
  }
  \caption{\edit{Quantitative comparisons of a), c) and e) reconstruction accuracy in Chamfer distance, and b), d) and f) the distance field accuracy in RMSE for the Cow and Lady (top), the Newer College (middle) and Mai City (bottom) datasets respectively.}}
  \label{all_plot_evaluation}
\end{figure}
\subsection{Reconstruction Accuracy}
For a fair comparison, we process every frame in each dataset for each framework.
VDBFusion and VDB-GPDF both produce similar meshes of the scene shown in Fig.~\ref{cow_mesh_vdbfusion} and Fig.~\ref{cow_mesh_ours}, respectively. 
VDB-GPDF produces a more complete and natural reconstruction, especially when the sensor is parallel to the observed surface. 
It can be seen that there is a gap at the top of the mattresses on the left-hand side in the VDBFusion result, whereas VDB-GPDF does not have the same gap. In addition, our colour is nicely fused with a sample of the RGB camera data for qualitative comparison. 

We compare the maps on both datasets quantitatively using the Chamfer distance metrics~\cite{mescheder2019occupancyChamfer} against VDB-Fusion and VDB-EDT. \edit{As demonstrated in Fig.~\ref{cow:chamfer} on the Cow and Lady,~\ref{newer:chamfer} Newer college and~\ref{mai:chamfer} Mai City,} our map quality outperforms VDB-EDT in different resolutions. Due to our voxelisation method being applied, we have comparable results to VDBFusion in reconstruction accuracy. 
In Fig.~\ref{teaser} on Newer College, we qualitatively show our mesh coloured by fused LiDAR intensity and the distance field output. Fig.~\ref{teaser1} shows the incrementally built dense reconstruction. We zoomed in Fig.~\ref{teaser2} to show the textures on the ceiling inside the corridor and Fig.~\ref{teaser3} the stairs and windows at the corner of the quad. \markred{Note that there exists a trade-off between accuracy and efficiency. For instance, efficiency decreases, and accuracy increases when more training points are added in L-GPDF and denser testing points are used in the frustum. In the results presented here, we achieve comparable reconstruction and distance accuracy with better time consumption.}
\begin{figure}[t]
  \centering
  \subfloat{\includegraphics[height=1.35cm]{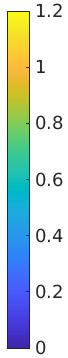}}
  \setcounter{subfigure}{0}  
  \subfloat[Ground Truth\label{distance_c:ground_truth}]{\includegraphics[height=1.35cm]{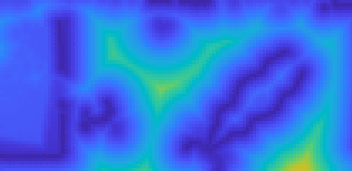}}
  \hfill
  \subfloat[VDB-EDT \label{distance_c:edt}]{\includegraphics[height=1.35cm]{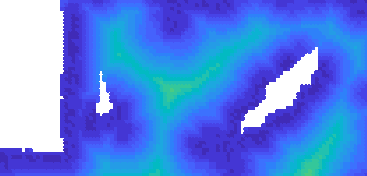}}
  \hfill
  \subfloat[Voxblox \label{distance_c:vox}]{\includegraphics[height=1.35cm]{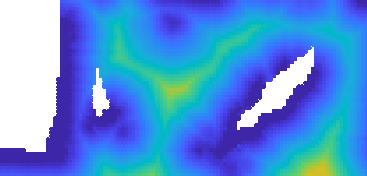}}\\
  \hfill
  \subfloat[w/o G-GPDF \label{distance_c:ours_direct_mode}]{\includegraphics[height=1.35cm]{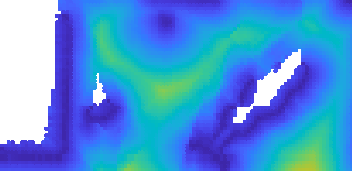}}
  \hfill
  \subfloat[with G-GPDF\label{distance_c:ours}]{\includegraphics[height=1.35cm]{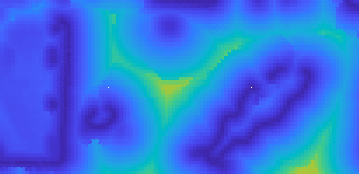}}
  \hfill
  \subfloat{\includegraphics[height=1.35cm]{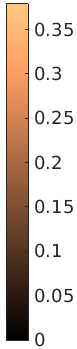}}
  \setcounter{subfigure}{5}  
  \subfloat[VDB-EDT error \label{distance_c:error_edt}]{\includegraphics[height=1.35cm]{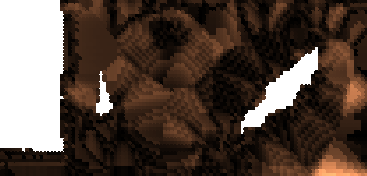}}
  \hfill
  \subfloat[\small{Voxblox error} \label{distance_c:error_vox}]{\includegraphics[height=1.4cm]{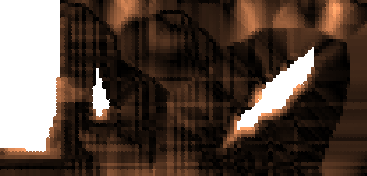}}
  \hfill
  \subfloat[w/o G-GPDF error\label{distance_c:error_ours_direct_mode}]{\includegraphics[height=1.4cm]{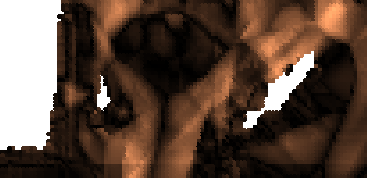}}
  \hfill 
  \subfloat[with G-GPDF error\label{distance_c:error_ours}]{\includegraphics[height=1.4cm]{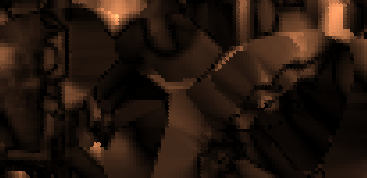}}
  \caption{Estimated EDF of cow and lady in e) is accurate and closely resembles the ground truth in a). Our VDB-GPDF with G-GPDF in i) produces lower and smoother errors than others.}
  \label{distance_cow_slice}
\end{figure}
\begin{figure}[t]
  \centering
  \resizebox{1.0\linewidth}{!}{
  \subfloat{\includegraphics[height=3.5cm]{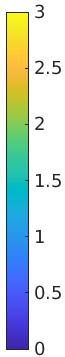}}
  \setcounter{subfigure}{0}  
  \subfloat[GT\label{distance_n:ground_truth}]{\includegraphics[height=3.5cm]{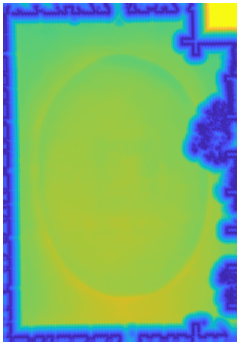}}
  \hfill
  \subfloat[VDB-EDT \label{distance_n:edt}]{\includegraphics[height=3.5cm]{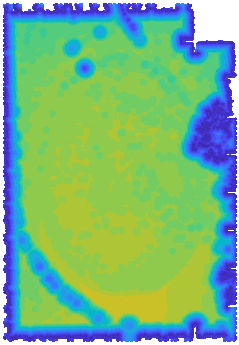}}
  \hfill
  \subfloat[Voxblox \label{distance_n:vox}]{\includegraphics[height=3.5cm]{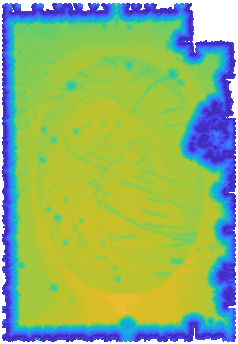}}
  \hfill
  \subfloat[VDB-GPDF \label{distance_n:ours}]{\includegraphics[height=3.5cm]{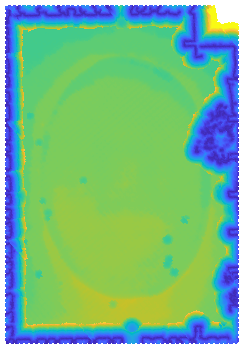}}
  }
  \resizebox{1.0\linewidth}{!}{
  \subfloat{\includegraphics[height=2cm]{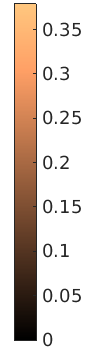}}
  \hfill
  \setcounter{subfigure}{4}  
  \subfloat[VDB-EDT error \label{distance_n:edt_error}]{\includegraphics[height=2cm]{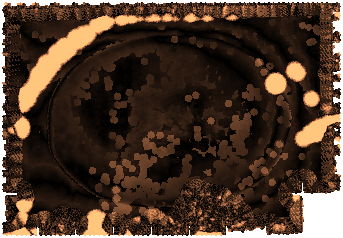}}
  \hfill
  \subfloat[Voxblox error \label{distance_n:vox_error}]{\includegraphics[height=2cm]{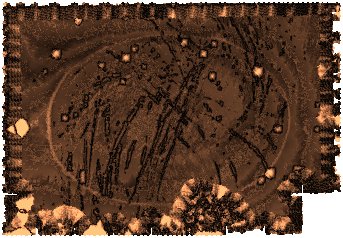}}
  \hfill
  \subfloat[Our error \label{distance_n:ours_error}]{\includegraphics[height=2cm]{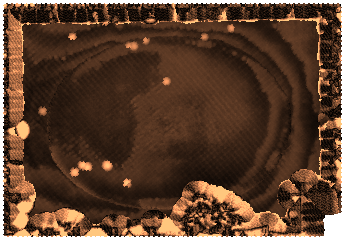}}
  }
  \caption{Estimated EDF of the newer college in d) is accurate and closely resembles the ground truth in a). b) fails to clean the areas where moving objects are present. c) does not produce enough smoothness in the field. Similarly, Our VDB-GPDF in h) produces lower values of distance errors.}
  \label{distance_newer_slice}
\end{figure}
\subsection{Distance Accuracy}
We evaluate the proposed GP distance field with VDB structure against Voxblox and VDB-EDT since the VDBFusion framework has no EDF. We use the ground truth point cloud to compute the distance field ground truth numerically. We compute the distance field on a regular 3D grid of 5cm resolution for the RGB-D case and 10 cm for the LiDAR case. The RMSE error given the ground truth distance field is shown \edit{in Fig.~\ref{cow:distance},~\ref{newer:distance} and ~\ref{mai:distance}.} Our distance field has the most accurate results. Voxblox produces the large distance RMSE for the \edit{Cow and Lady and Mai City datasets}, and VDB-EDT has a larger error on the Newer College dataset than others due to failure to clear the moving objects successfully.

To further examine the behaviour of VDB-GPDF, we compare 2D slices of the ground-truth and estimated distance fields for the Cow and Lady in Fig.~\ref{distance_cow_slice}.
We choose a horizontal slice $0.9$m above the ground, roughly cutting the scene objects in the middle. In addition, we evaluate the EDF with and without the G-GPDF enabled to show that G-GPDF produces more accurate distance inference than fusing the ESDF directly. Moreover, G-GPDF is a generative model that can generate distance and gradient inference even in places with no observations.
As shown in Fig.~\ref{distance_c:edt}, ~\ref{distance_c:vox} and Fig.~\ref{distance_c:ours_direct_mode}, VDB-EDT and Voxblox and VDB-GPDF without G-GPDF only compute the EDF within the sensor's field of view, whereas VDB-GPDF with G-GPDF naturally predicts the EDF value at all points as can be seen in Fig.~\ref{distance_c:ours}.
For a quantitative evaluation, we illustrate the distance error with a colour map in Fig.~\ref{distance_c:error_edt},~\ref{distance_c:error_vox},~\ref{distance_c:error_ours_direct_mode} and~\ref{distance_c:error_ours}. Our VDB-GPDF produces lower and smoother values of distance errors. Similarly, as shown in Fig.~\ref{distance_n:edt},~\ref{distance_n:vox} and~\ref{distance_n:ours}, our VDB-GPDF computes the similar EDF as the ground truth in Fig.~\ref{distance_n:ground_truth}.
For a quantitative evaluation, we plot the same distance absolute error in Fig.~\ref{distance_n:edt_error},~\ref{distance_n:vox_error} and~\ref{distance_n:ours_error}. Our VDB-GPDF generates lower and more consistent distance error values.

\subsection{Dynamic Performance}
\begin{figure}[ht]
  \centering
  \resizebox{1.0\linewidth}{!}{
  \subfloat[\label{kitti_1}]{\includegraphics[height=4cm]{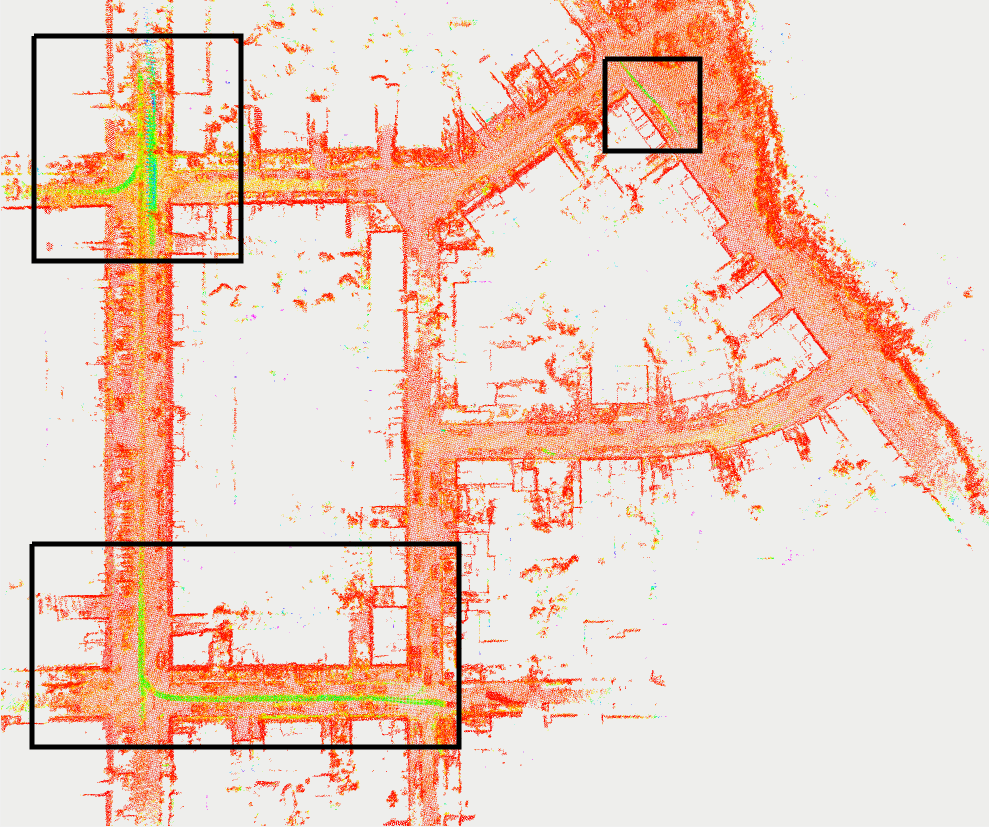}}
  \subfloat[\label{kitti_2}]{\includegraphics[height=4cm]{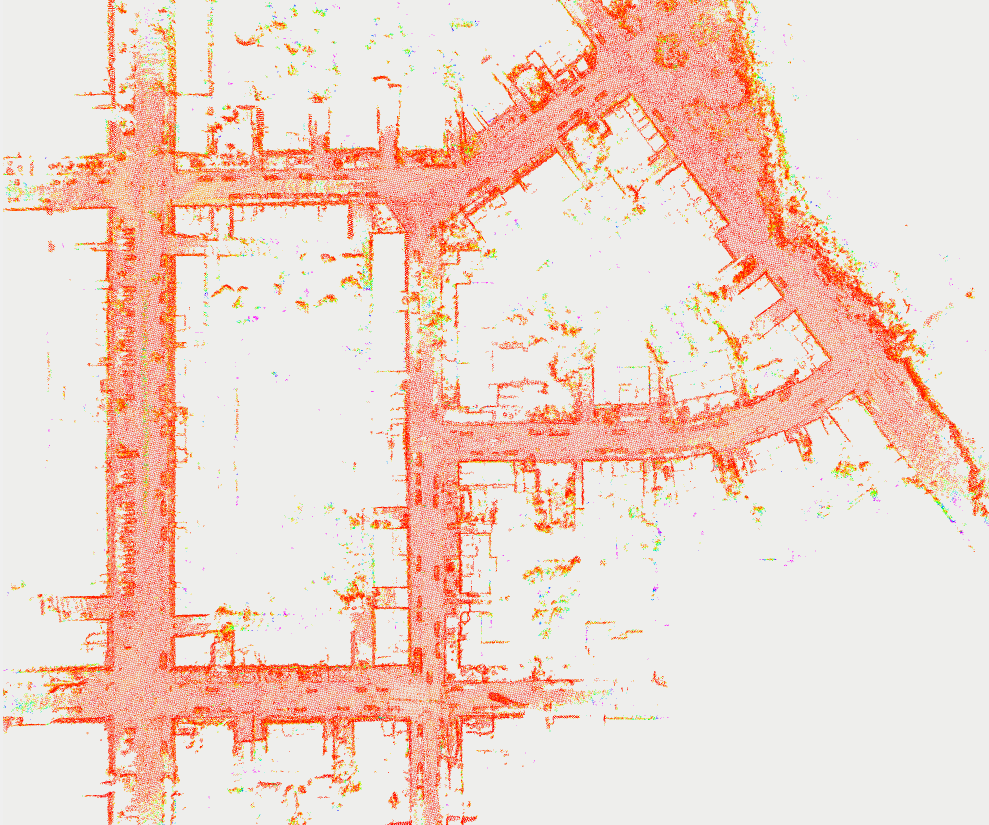}}
  }
  \caption{\markred{Kitti dataset map coloured by error  (red $\downarrow$ and green $\uparrow$) a) without space carving and b) with space carving.}}
  \label{kitti_figure}
  \vspace{-3ex}
\end{figure}

\edit{We evaluate our framework with and without the free space carving method on the large-scale KITTI dataset that contains multiple dynamic objects. \markred{We compare against the ground truth static map generated by~\cite{zhang2023dynamic} and colour the vertices of the reconstructed mesh with error.} The black sections in Fig.~\ref{kitti_1} indicate the map has accumulated defects due to the moving cars and pedestrians. The dynamic objects have been updated in the scene as shown in Fig.~\ref{kitti_2}.}

\section{Conclusion}\label{sec:conclusion}
We propose VDB-GPDF, a framework that couples VDB structure and GPDF for online incremental mapping.
It has the scalability to build a large-scale scene representation and is compatible with different types of depth sensors. 
It offers a comparable reconstruction of the scene environment and accurate Euclidean distance against state-of-the-art approaches.
VDB-GPDF can implicitly adapt to dynamic
changes and probabilistically handle noisy measurements.
VDB-GPDF provides flexible multiple outputs such as Euclidean distance and gradient fields, surface properties including colour and intensity, and a dense mesh with informative textures. 
Future work considers unlocking the ability to predict motion using the efficiently built changing distance field.

\bibliographystyle{IEEEtran}
\bibliography{reference}

\begin{thebibliography}{10}
\providecommand{\url}[1]{#1}
\csname url@samestyle\endcsname
\providecommand{\newblock}{\relax}
\providecommand{\bibinfo}[2]{#2}
\providecommand{\BIBentrySTDinterwordspacing}{\spaceskip=0pt\relax}
\providecommand{\BIBentryALTinterwordstretchfactor}{4}
\providecommand{\BIBentryALTinterwordspacing}{\spaceskip=\fontdimen2\font plus
\BIBentryALTinterwordstretchfactor\fontdimen3\font minus \fontdimen4\font\relax}
\providecommand{\BIBforeignlanguage}[2]{{%
\expandafter\ifx\csname l@#1\endcsname\relax
\typeout{** WARNING: IEEEtran.bst: No hyphenation pattern has been}%
\typeout{** loaded for the language `#1'. Using the pattern for}%
\typeout{** the default language instead.}%
\else
\language=\csname l@#1\endcsname
\fi
#2}}
\providecommand{\BIBdecl}{\relax}
\BIBdecl

\bibitem{oleynikova_voxblox_2017}
H.~Oleynikova, Z.~Taylor, M.~Fehr, R.~Siegwart, and J.~Nieto, ``Voxblox: {Incremental} {3D} {Euclidean} {Signed} {Distance} {Fields} for on-board {MAV} planning,'' in \emph{2017 {IEEE}/{RSJ} {International} {Conference} on {Intelligent} {Robots} and {Systems} ({IROS})}.\hskip 1em plus 0.5em minus 0.4em\relax IEEE, Sep. 2017.

\bibitem{han2019fiesta}
L.~Han, F.~Gao, B.~Zhou, and S.~Shen, ``Fiesta: Fast incremental euclidean distance fields for online motion planning of aerial robots,'' in \emph{2019 IEEE/RSJ IROS}.\hskip 1em plus 0.5em minus 0.4em\relax IEEE, 2019, pp. 4423--4430.

\bibitem{vizzo2022vdbfusion}
I.~Vizzo, T.~Guadagnino, J.~Behley, and C.~Stachniss, ``Vdbfusion: Flexible and efficient tsdf integration of range sensor data,'' \emph{Sensors}, vol.~22, no.~3, p. 1296, 2022.

\bibitem{zhu_vdb-edt_2021}
\BIBentryALTinterwordspacing
D.~Zhu, C.~Wang, W.~Wang, R.~Garg, S.~Scherer, and M.~Q.-H. Meng, ``\BIBforeignlanguage{en}{{VDB}-{EDT}: {An} {Efficient} {Euclidean} {Distance} {Transform} {Algorithm} {Based} on {VDB} {Data} {Structure}},'' May 2021. [Online]. Available: \url{https://arxiv.org/abs/2105.04419v1}
\BIBentrySTDinterwordspacing

\bibitem{bai_vdbblox_2023}
Y.~Bai, Z.~Miao, X.~Wang, Y.~Liu, H.~Wang, and Y.~Wang, ``Vdbblox: Accurate and efficient distance fields for path planning and mesh reconstruction,'' in \emph{2023 IEEE/RSJ International Conference on Intelligent Robots and Systems (IROS)}.\hskip 1em plus 0.5em minus 0.4em\relax IEEE, 2023, pp. 7187--7194.

\bibitem{wu_faithful_2021}
L.~Wu, K.~M.~B. Lee, L.~Liu, and T.~Vidal-Calleja, ``Faithful euclidean distance field from log-gaussian process implicit surfaces,'' \emph{IEEE Robotics and Automation Letters}, vol.~6, no.~2, pp. 2461--2468, 2021.

\bibitem{wu_log-gpis-mop_2023}
L.~Wu, K.~M.~B. Lee, C.~Le~Gentil, and T.~Vidal-Calleja, ``Log-{GPIS}-{MOP}: {A} {Unified} {Representation} for {Mapping}, {Odometry}, and {Planning},'' \emph{IEEE Transactions on Robotics}, pp. 1--17, 2023.

\bibitem{gentil_accurate_2023}
C.~Le~Gentil, O.-L. Ouabi, L.~Wu, C.~Pradalier, and T.~Vidal-Calleja, ``Accurate gaussian-process-based distance fields with applications to echolocation and mapping,'' \emph{IEEE Robotics and Automation Letters}, 2023.

\bibitem{bhoram_online_2019}
B.~Lee, C.~Zhang, Z.~Huang, and D.~D. Lee, ``Online continuous mapping using gaussian process implicit surfaces,'' in \emph{2019 International Conference on Robotics and Automation (ICRA)}.\hskip 1em plus 0.5em minus 0.4em\relax IEEE, 2019.

\bibitem{ali2024interactive}
U.~Ali, L.~Wu, A.~Mueller, F.~Sukkar, T.~Kaupp, and T.~Vidal-Calleja, ``Interactive distance field mapping and planning to enable human-robot collaboration,'' \emph{arXiv preprint arXiv:2403.09988}, 2024.

\bibitem{museth2013vdb}
K.~Museth, ``Vdb: High-resolution sparse volumes with dynamic topology,'' \emph{ACM transactions on graphics (TOG)}, 2013.

\bibitem{pan_voxfield_2022}
Y.~Pan, Y.~Kompis, L.~Bartolomei, R.~Mascaro, C.~Stachniss, and M.~Chli, ``Voxfield: {Non}-{Projective} {Signed} {Distance} {Fields} for {Online} {Planning} and {3D} {Reconstruction},'' in \emph{2022 {International} {Conference} on {Intelligent} {Robots} and {Systems} ({IROS})}, 2022, pp. 5331--5338.

\bibitem{museth2013openvdb}
K.~Museth, J.~Lait, J.~Johanson, J.~Budsberg, R.~Henderson, M.~Alden, P.~Cucka, D.~Hill, and A.~Pearce, ``Openvdb: an open-source data structure and toolkit for high-resolution volumes,'' in \emph{Acm siggraph 2013 courses}, 2013, pp. 1--1.

\bibitem{goel2024distance}
K.~Goel and W.~Tabib, ``Distance and collision probability estimation from gaussian surface models,'' \emph{arXiv preprint arXiv:2402.00186}, 2024.

\bibitem{park_deepsdf_2019}
J.~J. Park, P.~Florence, J.~Straub, R.~Newcombe, and S.~Lovegrove, ``Deepsdf: Learning continuous signed distance functions for shape representation,'' in \emph{Proceedings of the IEEE/CVF conference on computer vision and pattern recognition}, 2019, pp. 165--174.

\bibitem{gropp_learning_shapes_2020}
A.~Gropp, L.~Yariv, N.~Haim, M.~Atzmon, and Y.~Lipman, ``Implicit geometric regularization for learning shapes,'' \emph{arXiv preprint arXiv:2002.10099}, 2020.

\bibitem{pantic_NeRF_planning_2022}
M.~Pantic, C.~Cadena, R.~Siegwart, and L.~Ott, ``Sampling-free obstacle gradients and reactive planning in neural radiance fields (nerf),'' \emph{arXiv preprint arXiv:2205.01389}, 2022.

\bibitem{ortiz_isdf_2022}
J.~Ortiz, A.~Clegg, J.~Dong, E.~Sucar, D.~Novotny, M.~Zollhoefer, and M.~Mukadam, ``isdf: Real-time neural signed distance fields for robot perception,'' \emph{arXiv preprint arXiv:2204.02296}, 2022.

\bibitem{huang_di_fusion_2021}
J.~Huang, S.-S. Huang, H.~Song, and S.-M. Hu, ``Di-fusion: Online implicit 3d reconstruction with deep priors,'' in \emph{Proceedings of the IEEE/CVF Conference on Computer Vision and Pattern Recognition}, 2021, pp. 8932--8941.

\bibitem{vasilopoulos_hio_sdf_2023}
V.~Vasilopoulos, S.~Garg, J.~Huh, B.~Lee, and V.~Isler, ``Hio-sdf: Hierarchical incremental online signed distance fields,'' \emph{arXiv preprint arXiv:2310.09463}, 2023.

\bibitem{GPbook}
C.~E. Rasmussen and C.~K. Williams, \emph{Gaussian Processes for Machine Learning}.\hskip 1em plus 0.5em minus 0.4em\relax Cambridge, Mass.: MIT Press, 2006.

\bibitem{marching}
W.~E. Lorensen and H.~E. Cline, ``Marching cubes: A high resolution 3d surface construction algorithm,'' in \emph{Proceedings of the 14th Annual Conference on Computer Graphics and Interactive Techniques}, ser. SIGGRAPH '87.\hskip 1em plus 0.5em minus 0.4em\relax Association for Computing Machinery, 1987.

\bibitem{tsdf}
B.~Curless and M.~Levoy, ``A volumetric method for building complex models from range images,'' in \emph{Proceedings of the 23rd Annual Conference on Computer Graphics and Interactive Techniques}, 1996.

\bibitem{ramezani2020newer}
M.~Ramezani, Y.~Wang, M.~Camurri, D.~Wisth, M.~Mattamala, and M.~Fallon, ``The newer college dataset: Handheld lidar, inertial and vision with ground truth,'' in \emph{2020 IEEE/RSJ International Conference on Intelligent Robots and Systems (IROS)}, 2020, pp. 4353--4360.

\bibitem{vizzo2021poisson}
I.~Vizzo, X.~Chen, N.~Chebrolu, J.~Behley, and C.~Stachniss, ``Poisson surface reconstruction for lidar odometry and mapping,'' in \emph{2021 IEEE international conference on robotics and automation (ICRA)}.

\bibitem{behley2019semantickitti}
J.~Behley, M.~Garbade, A.~Milioto, J.~Quenzel, S.~Behnke, C.~Stachniss, and J.~Gall, ``Semantickitti: A dataset for semantic scene understanding of lidar sequences,'' in \emph{Proceedings of the IEEE/CVF international conference on computer vision}, 2019, pp. 9297--9307.

\bibitem{mescheder2019occupancyChamfer}
L.~Mescheder, M.~Oechsle, M.~Niemeyer, S.~Nowozin, and A.~Geiger, ``Occupancy networks: Learning 3d reconstruction in function space,'' in \emph{Proceedings of the IEEE/CVF conference on computer vision and pattern recognition}, 2019, pp. 4460--4470.

\bibitem{zhang2023dynamic}
Q.~Zhang, D.~Duberg, R.~Geng, M.~Jia, L.~Wang, and P.~Jensfelt, ``A dynamic points removal benchmark in point cloud maps,'' in \emph{2023 IEEE 26th International Conference on Intelligent Transportation Systems (ITSC)}.\hskip 1em plus 0.5em minus 0.4em\relax IEEE, 2023, pp. 608--614.

\end{thebibliography}

\end{document}